# Design of the Propulsion System of Nano satellite: StudSat2

[1] Roshan Sah, [2] Prabin Sherpaili, [3] Apurva Anand, [4] Sandesh Hegde

[1] IIT Kharagpur, [2], [3] NMIT, [4] Concordia University

*Abstract:* The increase in the application of the satellite has skyrocketed the number of satellites especially in the low earth orbit (LEO). The major concern today is after the end–of–life, these satellites become debris which negatively affects the space environment. As per the International guidelines of the European Space Agency, it is mandatory to deorbit the satellite within 25years of end-of-life. StudSat-1, which was successfully launched on 12th July 2010, is the first Pico satellite developed in India by undergraduate students from seven different engineering colleges across South India. Now, the team is developing StudSat-2, which is India's first twin satellite mission having two Nano satellites whose overall mass is less than 10kg. This paper is aimed to design the Propulsion system, Cold Gas thruster, to deorbit StudSat-2 from its original orbit to lower orbit (600km to 400km). The propulsion system mainly consists of a storage tank, pipes, Convergent-Divergent nozzle, and electronic actuators. The paper also gives information about the components of cold gas thruster, which have been designed in the CATIA V5, and the structural and flow analysis of the same has been done in ANSYS. The concept of Hohmann transfer has been used to deorbit the satellite and STK has been used to simulate it.

*Index terms--* debris, de-orbit, nozzle, propellant, thruster.

## I. INTRODUCTION

The satellites orbiting the earth are of no use after the end of their life. This necessitates the de-orbiting of satellite so that the functional satellites are in no danger of collision with the non-functional satellites. Moreover, it becomes essential to ensure that the de-orbiting takes place in lesser time to reduce the probability of collision. Hence, the need of an active method to de-orbit the satellite arises.

### A. Objective:
To design a cold gas propulsion system to de-orbit Nanosatellite.

### B. Abbreviations and Acronyms:

$V_t$ = Velocity in transfer orbit in m/s,
$\mu$ = Gravitational parameter for earth,
$R$ = Radius of orbit,
$\Delta V$ = Change in velocity,
$\varepsilon$ orbit = Specific mechanical energy in orbit,
TOF = Time of flight,
$P_c$ = Chamber pressure,
$P_e$ = Exit pressure,
$M_e$ = Exit Mach number,
$T$ = Temperature,
$A_t$ = Throat area.

## II. LITERATURE SURVEY

Orbit transfer is a common process in astrodynamics and Hohmann transfer is used as it is the most effective method. Chemical propellants (solid or liquid) are used for Inter-planetary maneuver like mars missions etc. After many catastrophes caused by the space debris, major focus has been given to re-orbit and de-orbit of satellite. For the satellites in the lower earth orbit, de-orbiting is preferred and for the satellites in the higher orbits, re-orbiting to Grave Yard orbit is preferred.

Deorbiting of satellite is new and rule for deorbit of small satellite have not been fully implemented. Various space agencies like NASA, ESA and universities like UTIAS SFL, Surrey space center, JPL Caltech, DLR (German Aerospace Center) Braunschweig, University of Patras etc are working in deorbit of satellite. UTIAS SFL had successfully lunched canX series in which cold gas thruster is used for station keeping and currently working to develop cold gas thruster for deorbiting.

## III. METHODOLOGY

Conceptual design was done based on the volume available in the satellite for the thruster and the required thrust. The 3D modeling was done in CATIA V5, which included propellant tank, pipes, nozzle and the valves. Structural and flow analysis was done in ANSYS Static Structure and CFX. The mesh was refined using O-grid to capture the boundary phenomena. The skewness for the mesh was 0.7 to 1, where 0 is considered to be the worst and 1 is ideal. The deorbit was simulated in STK.

## IV. THEORY BEHIND DE-ORBITING

Hohmann transfer orbit is used for de-orbiting as it is considered to be the simplest and the most efficient method of transfer a satellite in coplanar orbits and co-apsidal axis. It is a two-impulse elliptical transfer between two co-planar circular orbits. The transfer itself consists of an elliptical orbit with a perigee at the inner orbit and an apogee at the outer orbit. The mission is achieved by first transferring in the transfer orbit whose apogee is 600km and perigee is 400km. The transfer will be accomplished by firing the thruster. When the satellite perigee again thruster is fired again to transfer it in the final circular orbit of 400km.

Fig (1) shows the Hohmann transfer orbit with the direction of net velocity after firing the propulsion unit.

## V. COLD GAS THRUSTER

It works on the principle of conservation of energy, where the pressure energy of the gas is converted into kinetic energy by the use of nozzle providing thrust to de-orbit. A cold gas system consists of a pressurized tank containing propellant, nozzle, valves, pressure gauge and plumbing connecting them. Any gas can be used as a propellant. However, the gases with higher atomic mass is desirableas per Newton's Third Law, such as Heliu, Nitrogen, Xenon, etc.

Fig(2) shows the schematic representation of the thruster. To estimate the quantity of propellant, the pressure and temperature inside the tank, instrumentation devices are present. The valves control the release of propellant, and the nozzles accelerate the propellant to generate the desirable thrust. This system doesn't generate any net charge on the system nor does it contribute to any temperature rise, so known as Cold Gas.

*A. Components of Thruster*
*1) Storage Tank:*
Air is stored at a very high pressure in the tank. Expansion of air takes place from a higher pressure to space atmosphere where the pressure is negligible through convergent divergent nozzle. Tungsten matrix reinforced with boron fiber has a tensile strength of 3-4GPa and density of 2.23g/cc.

*2) Pressure regulating valve:*
Valve is used to supply air at constant pressure to the nozzle. Pressure valve is controlled electronically based in time.

*3) Nozzle:*
Nozzle converts pressure energy into kinetic energy. In our case, we use De-Laval nozzle for supersonic flow.

*B. Choice of propellant*
*1) Air:*
Air is easily available and its molecular mass is acceptable. Usually, air is non-corrosive at room temperature provided the amount of water vapor present in it is very less. While using dry air, the problem of corrosion doesn't arise. It is very cheap and can be stored at very high pressure.

## VI. EQUATIONS

$V_{t1} = \sqrt{2((\mu R_{orbit1}) + \mathcal{E}_{transfer})}$ (1)
$V_{orbit1} = \sqrt{2((\mu R_{orbit1}) + \mathcal{E}_{orbit1})}$ (2)
$\mathcal{E}_{transfer} = -\mu/2a_{transfer}$ (3)
$V_{orbit2} = \sqrt{2((\mu R_{orbit2}) + \mathcal{E}_{orbit2})}$ (4)
$\Delta V_2 = |V_{transfer\ orbit} - V_{orbit2}|$ (5)
$\Delta V = \Delta V_1 + \Delta V_2$ (6)
$TOF = \pi\sqrt{a_{transfer}^3/\mu}$ (7)
$T_c/T_e = 1 + (\gamma-1)*M_e^2/2$ (8)
$\Delta V = u_{eq}*\ln(m_i/m_f)$ (9)
$P_c = (b/r_i^2) - a$ (10)
$P_o = (b/r_e^2) - a$ (11)
$f = (b/r_i^2) + a$ (12)

## VII. DESIGN

Conceptual design is done based on the thrust required to de-orbit the satellite. Parameters like the volume of propellant, size of tank, pressure and temperature inside the tank, dimensions of supersonic nozzle and diameter of pipe are all designed based on the calculations performed using the equations 1-9 and 3D modeling of all the components are done using CATIA V5.

Fig (4) shows the different components of the satellites including the position of thruster inside the satellite.Table I, II and III include all the values required parameters obtained from calculations and the same dimensions are used for modelling.

## VIII. ANALYSIS

The values obtained by mathematical calculations are verified using MATLAB and the important parameters like exit velocity of air from nozzle, maximum hoop stress in the tank, maximum deformation of tank under the influence of internal pressure are simulates using ANSYS. Fig 5 to 12 show all the analysis results, including structural analysis as well as flow analysis.

## IX. DE-ORBITING TRAJECTORY

Systems Tool Kit (STK) is used to simulate the deorbit process using the Hohmann transfer method. STK gives the orbital parameters latitude, longitude and altitude during the deorbit period. The position of the satellite after every point of time is determined while de-orbiting of the satellite from 600km to 400km.

The graph 5 shows the variation of orbital parameters (lat, long and alt) during the deorbit. In the graph, the blue lines shows drop of altitude when the cold gas is fire in two stage. The variation is the final altitude is due to the lace of convergence and high tolerance in the tool stk. The main reason of using STK is the reliability of results as it is used by NASA and other organization for various space maneuvers simulation.

## X. COLLISION AVOIDANCE

Deorbiting of the satellite is an elegant and promising solution for space debris problem. But during the deorbiting there is the probability of collision of deorbiting satellite with the functional satellite which again creates serious problem. Several simulation performed shows that odds of collision is very low but for the better future of deorbiting collision avoidance study is important. Active method like thruster to avoid the collision are not applicable for the small satellite like STUDSAT-2 due to size, mass and other constrains.

If trajectory of the deorbiting satellite and the orbit of the functional satellite intersect and deorbiting satellite and functional are present in that point at same time, collision occurs. Realizing this, extensive study on the orbital dynamics and TLE was done. TLE is Two Line Element which contain the information about the revolving body like satellite. From TLE we can find the exact position of the satellite and then deorbit maneuver is started such that the deorbiting satellite.

## XI. RESULTS

From the static structure analysis of the tank in ANSYS workbench, the maximum equivalent Von-Mises obtained was 1.19Gpa which is lower than permissible safe load with load factor of 1.5 as shown in fig(6). As shown in Fig, the stress concentration is near the base periphery of tank. Similarly the total deformation of the tank is shown in fig(7). The maximum total deformation of satellite at center of the base of tank which is 2mm which is in acceptance range. The result of CFD is in close match with the calculation done. Various contour like pressure, temperature, Mach number are obtained from the CFX post from fig(10) to fig(12). The variation of Mach number, pressure and temperature at the central line along the length of the nozzle is shown in graph (2) to (4). Similarly to study the effect of the different turbulence model like Shear Stress Transport, BSL Reynold stress, SSG Reynold stress on the Mach number, temperature and pressure same C-D nozzle at same boundary condition was simulate for these turbulence model. The result is shown in the graphs.

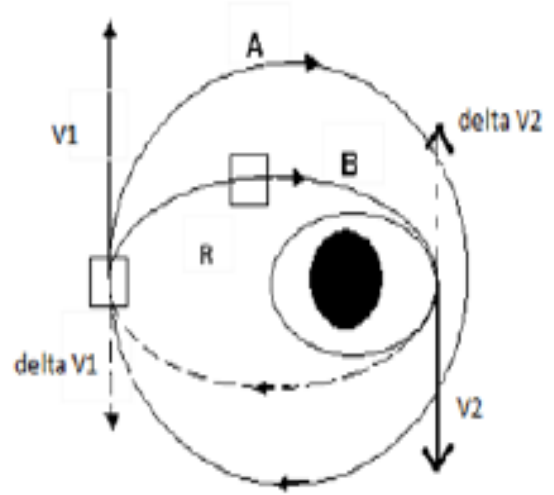

*Fig 1: Hohmann Transfer Orbits*

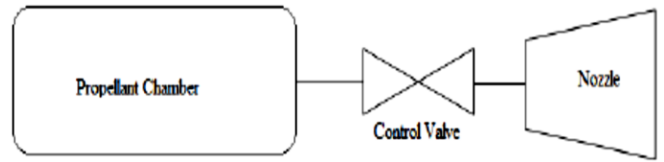

*Fig 2: Schematic Representation of Cold Gas Thruster*

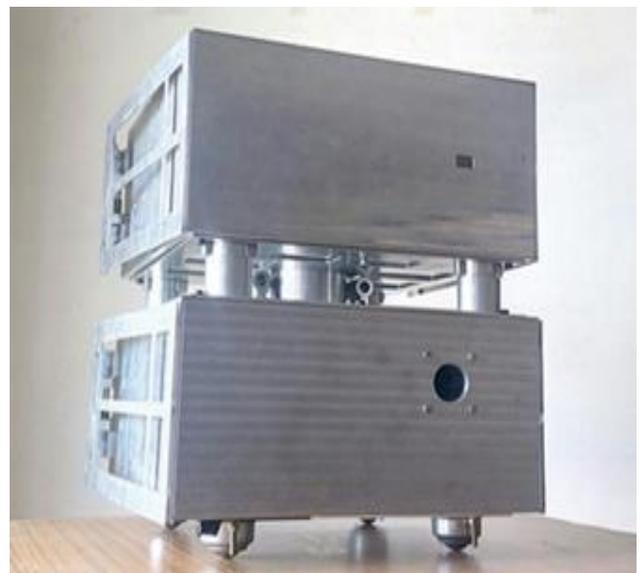

*Fig 3: StudSat-2*

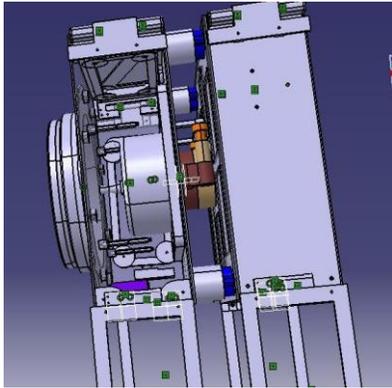
*Fig 4: Position of thruster in Satellite*

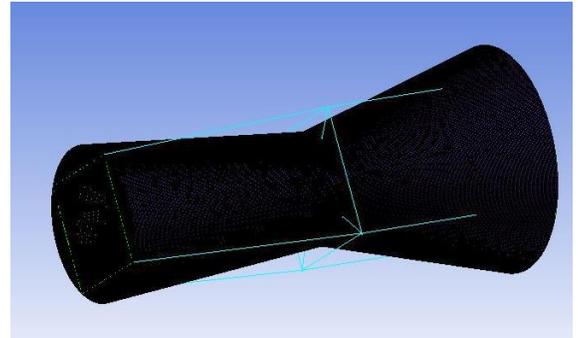
*Fig 8: Structured mesh of C-D Nozzle*

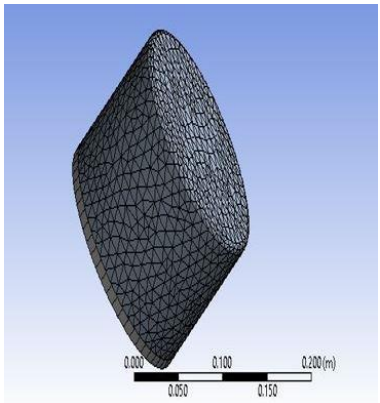
*Fig 5: mesh of tank*

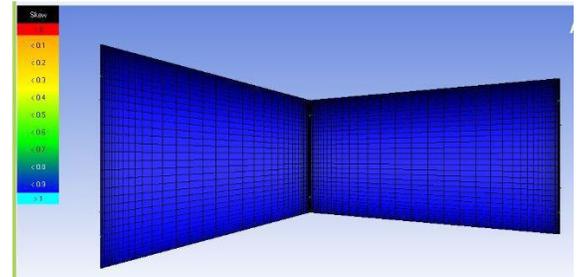
*Fig 9: mesh refinement*

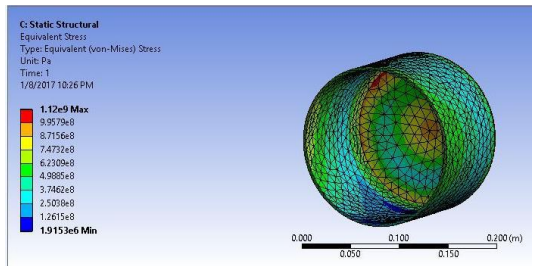
*Fig 6: Von-Mises Stress inside the tank*

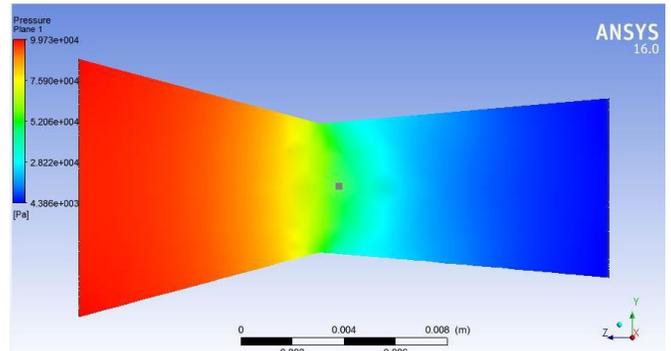
*Fig 10: Pressure Contour*

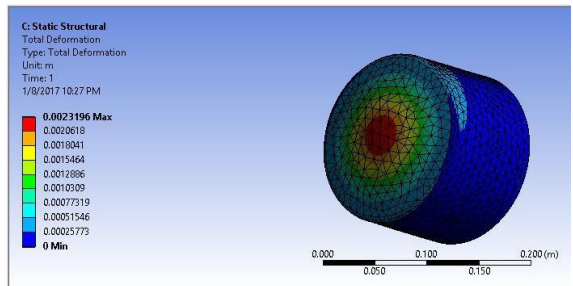
*Fig 7: Total Deformation of Pressurized Tank*

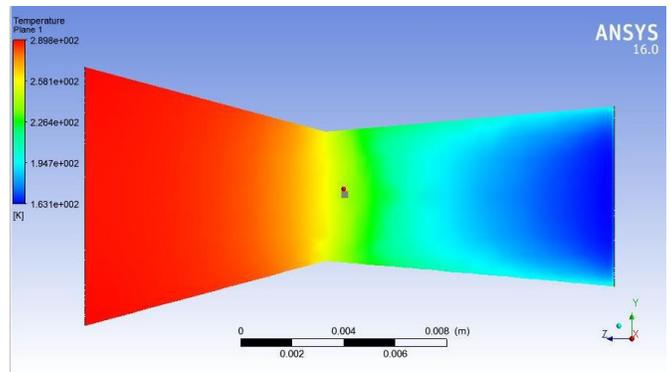
*Fig 11: Temperature Contour*

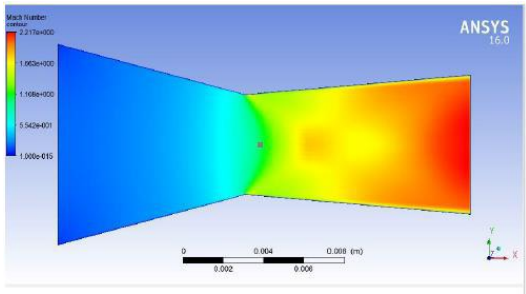
*Fig 12: Mach number Contour*

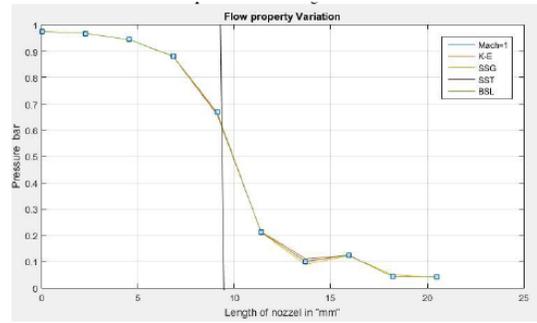
*Graph 3: Pressure vs length of nozzle*

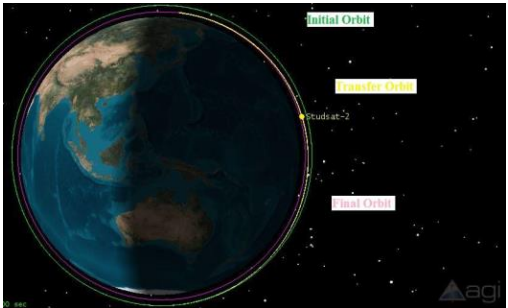
*Fig 13: different orbits of satellite*

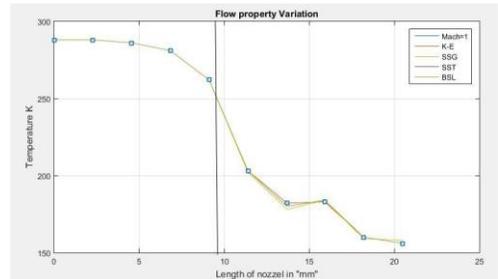
*Graph 4: Temperature vs length of nozzle*

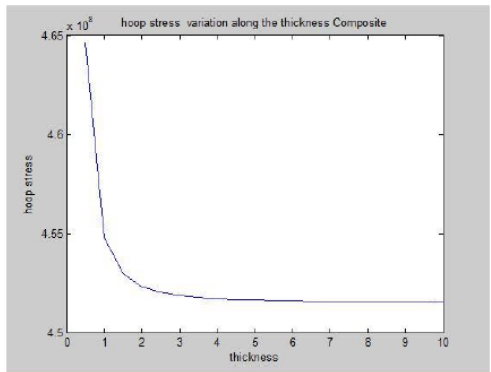
*Graph 1: Hoop Stress Variation Along thickness of Tank*

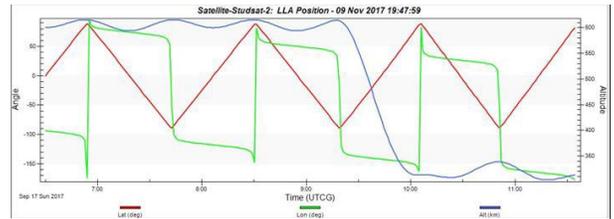
*Graph 5: Variation of orbital parameter during deorbit*

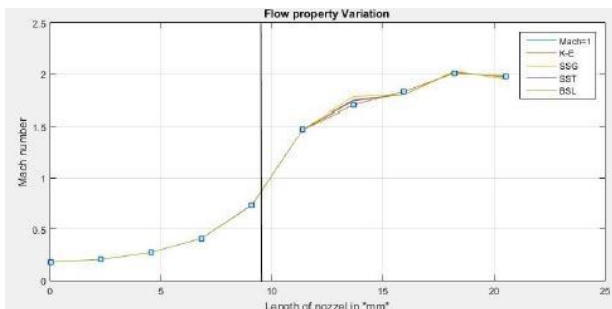
*Graph 2: Mach number vs length of nozzle*

*Table I: Orbital Parameters*

| Parameters | Calculated value for 600 to 400km | Unit |
|---|---|---|
| R earth | 6370 | Km |
| R orbit1 | 6970 | Km |
| R orbit2 | 6770 | Km |
| a transfer | 6870 | Km |
| ε transfer | -29.0102 | Km2/s2 |
| ε orbit1 | -28.5940 | Km2/s2 |
| ε orbit2 | -29.4387 | Km2/s2 |
| V t1 | 7.5070 | Km/s |
| V orbit1 | 7.5623 | Km/s |
| ΔV 1 | 0.05529 | Km/s |
| V t2 | 7.7288 | Km/s |
| V orbit2 | 7.6732 | Km/s |
| ΔV 2 | 0.05561 | Km/s |
| ΔV | 0.1109 | Km/s |
| TOF | 2833.5 | Sec |

*Table II: Nozzle Dimensions*

| Parameters | Values | Unit |
|---|---|---|
| Throat Area | 1.9635e-5 | $m^2$ |
| Exit diameter | 6.9559 | Mm |
| Inlet diameter | 10 | Mm |
| Nozzle half angle | 5 | degree |
| Length | 11.43- convergent 9.46-divergent | Mm |

*Table III: Tank dimensions*

| Parameters | Values | Unit |
|---|---|---|
| Inner diameter | 16.56 | Mm |
| Length | 12 | Mm |
| Thickness | 3.1 | Mm |

*Table IV: Boundary conditions and mesh description*

| Parameters | Values / description |
|---|---|
| Number of nodes | 122976 |
| Number of elements | 118275 |
| Number of hexahedrons | 118275 |
| Number of faces | 9158 |
| Reference pressure | 0.1bar |
| Inlet pressure | 1bar |
| Inlet temperature | 290K |
| Turbulence model | K-epsilon |

## XII. ACKNOWLEDGEMENT

We thank Dr. U. R. Rao, former Chairman of ISRO (Indian Space and Research Organization) for providing necessary guidance and support to us during this project. We express our gratitude to Dr. S Sandya and Dr. H.C. Nagaraj for encouragement and support.

## XIII. CONCLUSION

Despite adding to little complexity in the system, an active propulsive method such as cold gas thruster can considerably reduce de-orbiting time and easily avoid collision due to controlled thrust. Cold gas thruster can not only be used for small satellites, but it is equally effective in larger satellites, where temperature gradient is a major concern.